\documentclass{article}
\usepackage{microtype}
\usepackage{graphicx,subcaption}
\usepackage{natbib}
\usepackage{wrapfig}
\usepackage{booktabs} % for professional tables
\usepackage[T1]{fontenc}
\usepackage{dsfont}
\usepackage[ruled,vlined, noend]{algorithm2e}
\usepackage{amsmath, amsthm, amssymb, amsfonts}
\usepackage{array}
\usepackage{makecell, tabularx, multirow}
\usepackage{rotating}
\usepackage{diagbox}

\usepackage{shortbold}
\usepackage{hyperref}
\usepackage{xcolor}
\usepackage{enumitem}
\usepackage[title]{appendix}
\usepackage[capitalise]{cleveref}
\usepackage{arxiv}
\usepackage{ulem}
\usepackage{xr}
\usepackage{cancel}
\makeatletter

\author{
Asif Khan, Amos Storkey\\
School of Informatics, \\
University of Edinburgh, UK
  }

\title{ Contrastive Learning for Non-Local Graphs with Multi-Resolution Structural Views}
\begin{document}

\maketitle
\begin{abstract}
Learning node-level representations of heterophilic graphs is crucial for various applications, including fraudster detection and protein function prediction. In such graphs, nodes share structural similarity identified by the equivalence of their connectivity which is implicitly encoded in the form of higher-order hierarchical information in the graphs. The contrastive methods are popular choices for learning the representation of nodes in a graph. However, existing contrastive methods struggle to capture higher-order graph structures. To address this limitation, we propose a novel multiview contrastive learning approach that integrates diffusion filters on graphs. By incorporating multiple graph views as augmentations, our method captures the structural equivalence in heterophilic graphs, enabling the discovery of hidden relationships and similarities not apparent in traditional node representations. Our approach outperforms baselines on synthetic and real structural datasets, surpassing the best baseline by $16.06\%$ on Cornell, $3.27\%$ on Texas, and $8.04\%$ on Wisconsin. Additionally, it consistently achieves superior performance on proximal tasks, demonstrating its effectiveness in uncovering structural information and improving downstream applications.
\end{abstract}

\section{Introduction}
Graphs have emerged as prominent data structures for capturing complex relational interactions among entities in various domains, including social networks, molecular systems, and more~\citep{wang2022molecular, wang2021molclr}. 
The availability of diverse graph datasets~\citep{bojchevski2017deep, zitnik2017predicting, rozemberczki2021multi} and the need to address tasks such as molecular property prediction~\citep{wu2018moleculenet}, protein-protein interaction~\citep{zitnik2017predicting}, drug-disease associations~\citep{lin2020kgnn}, and inferring a new connection between entities~\citep{zhang2018link},
have fueled the development of machine learning (ML) approaches tailored for graph-structured data. 

Traditionally, supervised~\citep{cangea2018towards, shervashidze2011weisfeiler, dai2016discriminative, li2015gated}, semi-supervised~\citep{belkin2006manifold,weston2012deep, kipf2016semi, bui2018neural}, and unsupervised methods~\citep{kipf2016variational,belkin2003laplacian, cai2018comprehensive} have been employed in ML for graphs for addressing various problems. Supervised approaches rely on class labels to guide representation learning, but their reliance on annotated data poses challenges due to limited availability and high labelling costs. This limitation has motivated the exploration of self-supervised and unsupervised learning methods that can leverage the inherent structure of the data to learn meaningful representations without the need for explicit labels. Such approaches have gained significant attention due to their ability to efficiently learn representations in a data-driven manner while mitigating the annotation bottleneck.
\begin{figure}[t]
\includegraphics[width=0.62\columnwidth, trim={7cm 4cm 7cm 4cm}, clip]{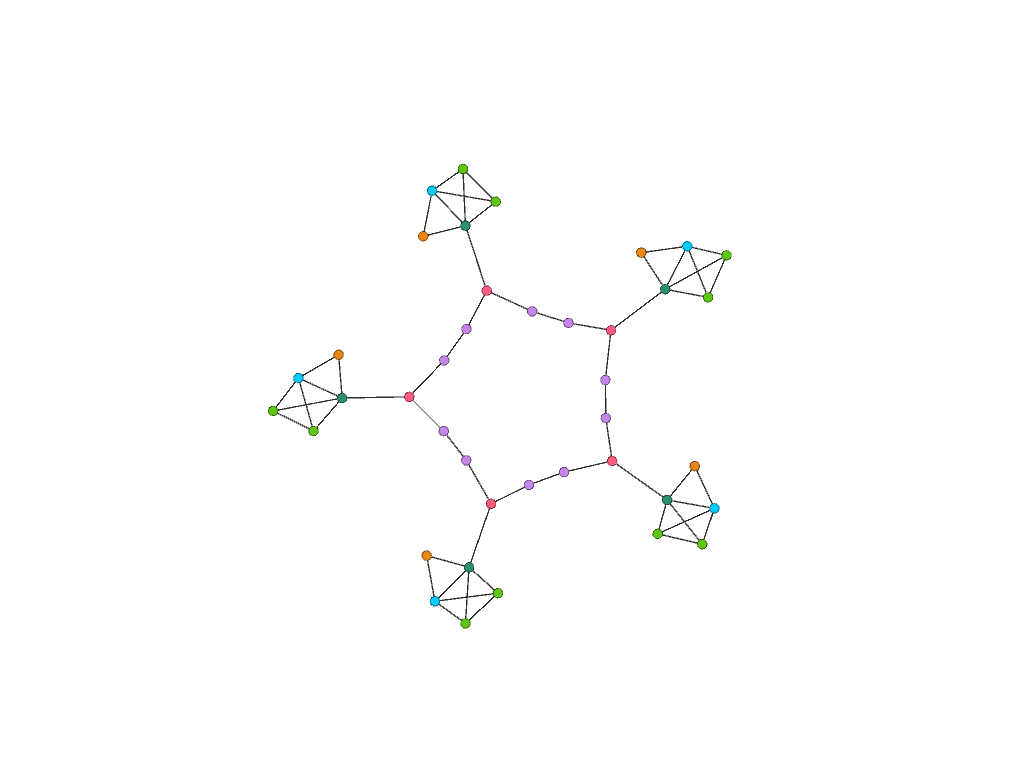}
\includegraphics[width=0.37\columnwidth]{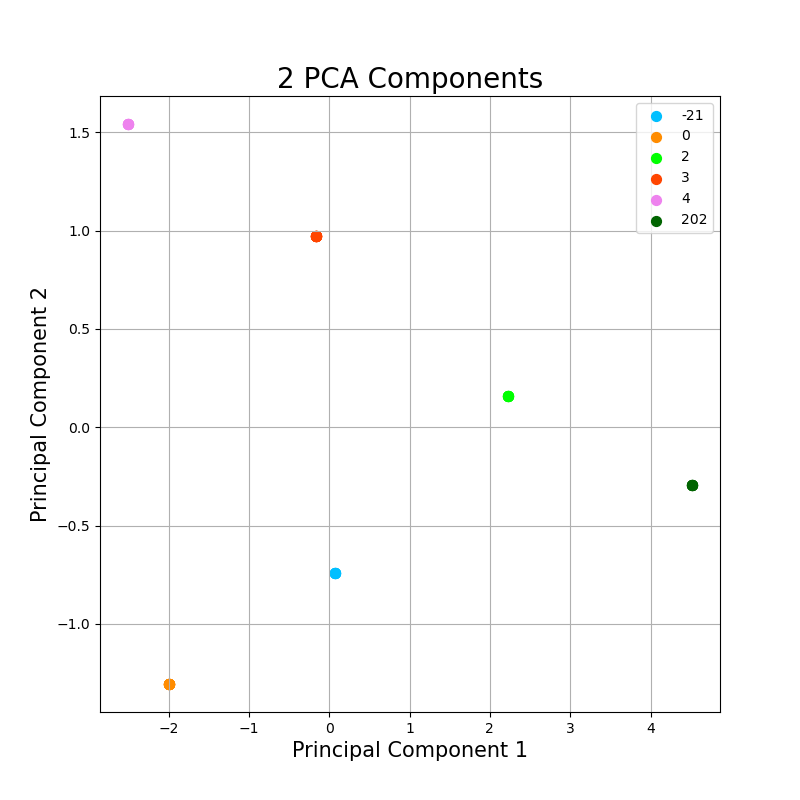}
\caption{An illustration of structural equivalence. 
The left panel depicts a toy graph, where nodes are assigned different colours to denote their roles within the graph structure. On the right panel, we present the first two principal components of the node embeddings obtained through our multi-resolution contrastive approach. These principal components are further colour-coded based on the node labels. Notably, the embeddings effectively capture the notion of structural similarity, as nodes with similar structures are assigned similar representations, as indicated by the consistent colouring of nodes with similar roles in the graph.}
\label{struct_eqv}
\end{figure}

Graph representation learning methods aim to capture the geometry of graph data by embedding nodes into a lower-dimensional space. The objective is to ensure that nodes with a particular notion of similarity in the graph are mapped to nearby locations in the embedding space. These learned embeddings can be readily used for various downstream tasks such as node classification, link prediction, and clustering~\citep{hamilton2020graph, chen2020graph, wang2017predictive}. However, it is crucial to incorporate appropriate assumptions during the training process to achieve better generalisation to downstream tasks. For instance, in a homophilic graph, where nodes with similar attributes tend to be connected, the representation space should prioritise the similarity of encodings between proximal nodes compared to non-proximal ones. Conversely, in a heterophilic graph, where structurally similar nodes are usually distantly connected, it is expected the embedding of these nodes in a representation space is to nearby locations. By tailoring the learning process to the specific characteristics of the graph, we can enhance the quality and applicability of the learned representations.

The contrastive approaches in the SSL paradigm utilise an objective to encourage similar samples (positives) to be nearby in the embedding space while dissimilar samples (negatives) are pushed farther apart~\citep{chen2017sampling, chen2020graph, jaiswal2020survey}. The positives are the alternative views of data obtained using augmentation transformations, and the negatives are sampled using a stochastic corruption strategy. SSL has been extended in graph-structured data to capture the inherent graph properties\citep{velickovic2019deep, you2020graph, hassani2020contrastive}. Graph contrastive learning (GraphCL;~\citealt*{you2020graph}) introduces task-specific augmentations to generate multiple graph views, enabling the learning of view-consistent node representations. Deep graph infomax (DGI;~\citealt*{velickovic2019deep}) maximises the mutual information between a node embedding and a global graph embedding obtained through a readout function. However, these methods have limitations in capturing global structural information due to their reliance on local graph networks represented by normalised adjacency matrices. To address this limitation, multi-view graph representation learning (MVGRL;~\citep{hassani2020contrastive}) utilises a random walk diffusion matrix as a global structural view of the graph. However, the dominance of low-frequency components in the random walk diffusion limits its ability to express higher-order hierarchical information crucial for capturing structural equivalence~\citep{gao2019geometric}.

The hierarchical nature of graphs comes with the higher-order relational information embedded within different levels of resolution~\citep{coifman2006diffusion}.In this paper, we argue that capturing such a hierarchy is vital for learning good representations, especially relevant for structural equivalence tasks. We propose a mechanism that utilises information from multi-resolution diffusion filters on graphs for representation learning. We achieve this by constructing a cascade of diffusion filters on the input graph, starting with a lazy diffusion operator and constructing coarser graphs using dilation filters. Each filter represents a different resolution of the diffusion operator applied to the graph. By training an encoder to map these coarse views to a feature space using a contrastive objective, we enable the encoder to learn representations invariant to multi-resolution views. This invariance accounts for the higher-order information necessary for identifying structural equivalence. Our empirical results demonstrate that our approach outperforms other SSL methods on various synthetic and real-world heterophilic graphs. By leveraging the principles of multi-view coding, our simple yet effective method holds promise for advancing representation learning on graph-structured data.

The choice of dilated filters offers several advantages. Each coarse view acts as a bandpass filter, capturing interdependency among nodes at varying resolutions. This approach provides augmentations of the input graph that explicitly presents the encoder network with a hierarchy of spatially localised information, enabling the learning of latent space that captures long-range dependencies, thereby accounting for structural equivalence. Furthermore, diffusion filters are stable, remaining robust to perturbations and deformations in the input node features of a graph. This stability is vital to the robustness of the learned representations, ensuring their reliability under real-world applications.
\section{Related Work}
\label{ch5:related}
Representation learning for graphs encompasses a wide range of approaches, including supervised and unsupervised methods. Several works have built on the success of contrastive learning in image data to propose algorithms for graph-structured data. In the context of this paper, we refer to the following works as particularly relevant contributions to representation learning for graphs.

\subsection{Random Walk Methods} 
Embedding models have gained significant popularity in natural language processing, as they enable the mapping of words to dense feature representations, where semantically similar words tend to be located close to each other in the embedding space. 
A notable example is Word2Vec, introduced by\citet{mikolov2013distributed}, which utilises the Skip-Gram model to maximise the log probability of a word given its context window. In doing so, Word2Vec effectively learns to map words with high co-occurrence within a context window to representations that are nearby under Euclidean distance. In the context of graph data, various embedding methods have employed random walk sequences to train models inspired by Word2Vec, such as DeepWalk~\citep{perozzi2014deepwalk}, node2vec\citep{grover2016node2vec}, subgraph2vec\citep{ narayanan2016subgraph2vec}, and many others. However, random walk-based methods have limitations in capturing long-term dependencies and often perform poorly on tasks requiring structural similarity.

To address the above challenges,~\citet{narayanan2016subgraph2vec} proposed an algorithm that learns node embeddings, such that the embeddings of nodes with a similar local structure are more similar. However, their algorithm relies on a predefined notion of local structure and may struggle with graphs exhibiting more complex structural equivalence. Several recent approaches~\citep{yanardag2015deep,al2019ddgk,borgwardt2020graph} have tackled the issue of capturing structural similarity by decomposing graphs into sub-structures and employing graph kernels to measure node similarity. Nonetheless, a fundamental limitation is that determining the appropriate sub-structures requires domain knowledge, which may not be readily available for a wide range of graph data.
\subsection{Spectral Methods}
Methods in this category rely on the spectral properties of adjacency or the Laplacian matrix of the graph. The eigenvectors of the Laplacian matrix are interpreted as frequencies in the Fourier space encoding information about graph properties, such as the size of cuts. Classical methods like Laplacian eigenmaps~\citep{belkin2003laplacian} use the eigenvectors corresponding to top-k eigenvalues of Laplacian as feature representations. However, top-k eigenvalues of Laplacian correspond to low-frequency components, they fail to capture higher frequencies important for long-term dependencies.

Alternatively, methods using the diffusion process are based on the heat kernel of a graph, which is the solution of the heat equation associated with the Laplacian operator of the graph. The heat kernel can be interpreted as a similarity matrix where each entry is an expected value of distance across all paths between pairs of nodes. The embeddings of the nodes are constructed by taking top-k eigenfunctions of the heat kernel matrix. Other approaches like ~\citet{tsitsulin2018verse} stack the trace of the diffusion matrix at different scales to get its feature representation which in turn is used for supervised classification. Still, under a large diffusion time, the contributions from higher frequency are suppressed, leading to ineffective representations for expressing higher-order structural information.
Several other approaches directly utilise spectral methods to learn filters that capture structural relationships, such as graph convolution (GCN)~\citep{kipf2016semi} ChebNet~\citep{defferrard2016convolutional}, CayleyNets~\citep{levie2018cayleynets}, and so on. These methods learn suitable filters based on the spectral properties of the graph to better capture structural relationships. 

While Laplacian-based methods and diffusion approaches have their merits, they are still limited in capturing multi-scale information crucial for complex structural relationships.
\begin{figure*}[ht!]
    \centering
    \includegraphics[height=7cm, width=0.9\columnwidth]{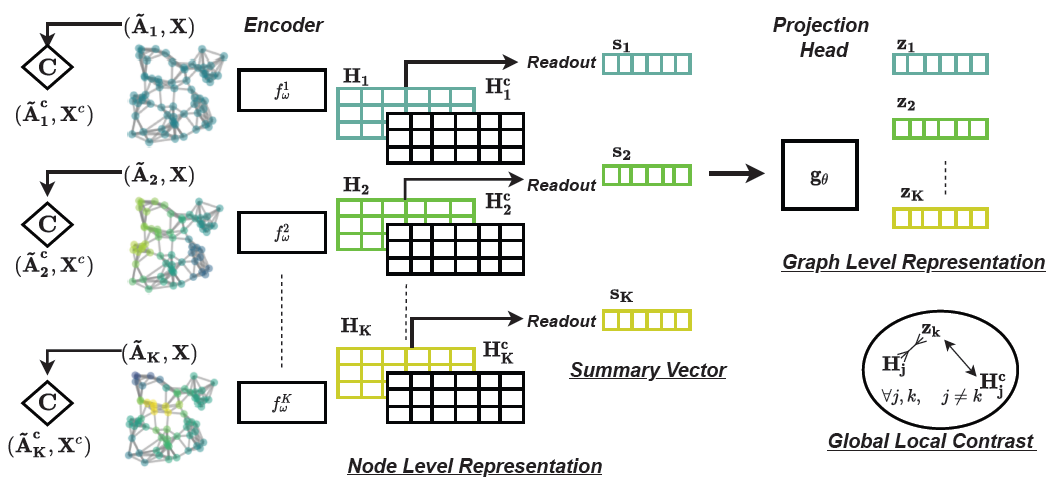}
    \caption{An illustration of our multi-resolution self-supervised learning pipeline. We first use diffusion wavelets to construct multi-resolution views $\{G_1,\cdots, G_K\}$ and combine them with attribute information to learn node level representation $\HBRV_k$ using an encoder network (we use GCN with two layers). For each resolution, we also generate a corrupted view using a stochastic corruption function $\mathbf{C}$ and map them to node-level representation $\HBRV_k^c$. Next, for each of the $k$ views, we pass the node representation matrix to a \textit{Readout} operation followed by an MLP to get the graph level representation $\zBRV_k$. A contrastive loss is trained to maximise consistency between node and graph representation of intra-views. The final node level representations are obtained by pooling across views as $\HBRV = \frac{\sum_{k} \HBRV_k}{K}$}.
    \label{fig:flowdiag}
\end{figure*}

\subsection{Contrastive methods} 
Contrastive approaches have emerged as a popular choice for representation learning. The idea is to learn representations that share information expressed in different data augmentations, a.k.a views. The methods hold application to a comprehensive set of problems encountered in computer-vision~\citep{chuang2020debiased, chen2020simple, he2020momentum}, reinforcement learning~\citep{laskin2020curl}, etc. The Skip-Gram training of DeepWalk is also a form of contrastive learning; it utilises positive and negative pairs of nodes based on a distance in a random walk and trains a discriminator to increase the score of positive samples and decrease the scores of negatives. More recent contrastive methods focus on different forms of discriminator function as well as ways for generating good negative samples~\citep{chen2017sampling, chen2020simple}.~\citet{velickovic2019deep} introduced deep graph infomax (DGI) for node representation learning that contrasts the local-global representations from the negative pairs. Another work, Infograph~\citep{sun2019infograph}, used infomax to learn graph-level representations.

A recent article closely related to our work is multi-view graph representation learning (MVGRL;~\citep{hassani2020contrastive}). MVGRL utilised multiple views of graphs obtained by varying time scales of the diffusion process and trained a neural network with the DGI objective. However, as noted by the authors, the results do not benefit from additional views. Our work differs in a critical aspect; we do not use a single global diffusion view; instead, we construct a cascade of filters that provide multi-resolution views of the structural information.

\subsection{Autoencoder methods} The encoder-decoder model and its probabilistic counterpart are widely used for representation learning. ~\citet{kipf2016variational} extended variational autoencoder (VAE)~\citep{kingma2013auto} to graph-structured data by employing layers of GCN as an encoder. They used a Gaussian prior on node representations and a dot product decoder to reconstruct the graph structure. Subsequent works proposed a more expressive GNN architecture for the encoder~\citep{hamilton2017inductive, grover2019graphite, pan2018adversarially, yang2020secure} or the decoder~\citep{wang2017mgae, park2019symmetric,shi2020effective}.  
However, most of the above methods focused only on local neighbourhood structures during the reconstruction phase, resulting in poor performance on structural equivalence tasks. To address this limitation~\citet{tang2022graph} proposed an approach called neighbourhood Wasserstein reconstruction (NWR). NWR introduces an objective function that increases the capacity of representation in capturing both proximity and structure information. Instead of using a probabilistic encoder, NWR training utilises a deterministic one. Our approach is not directly comparable to NWR as our method belongs to a self-supervised class that does not rely on a decoder neural network.
\subsection{Diffusion wavelets} Diffusion wavelets, introduced by~\citet{coifman2006diffusion}, offer a powerful yet simple tool for multi-resolution analysis of signals. The central idea is to start with a lazy diffusion operator and examine the difference in heat diffusion across varying timescales. Building on the work of~\citet{mallat2012group},~\citet{gama2018diffusion} utilised diffusion wavelets to construct multi-resolution filter banks on graphs. They demonstrated the stability of such filters in the face of signal deformation and perturbation, enabling the training of deep and stable graph neural networks. More recently,~\citet{gao2019geometric} showcased that diffusion wavelets could be used to learn a universal representation of graphs. While~\citet{tong2020data} proposed a method to learn the scale parameter of filters using stochastic softmax tricks~\citep{NEURIPS2020_3df80af5}. In contrast to these approaches, our work employs multi-resolution filters as alternate graph views; that, when combined with a contrastive objective, can learn resolution-invariant representations useful for structural equivalence in graphs.

Another wavelet-based method, GraphWave~\citep{donnat2018learning}, utilises wavelet diffusion to learn multi-resolution structural embedding. Our work differs from GraphWave in two aspects. Firstly, while GraphWave solely operates on structural information, our approach can leverage both structure and node attributes. Secondly, GraphWave learns embeddings by characterising wavelet filters in the spectral domain using the characteristic function of the Laplacian eigenvalues. In contrast, we construct dyadic scale views of the graph and utilise them as augmentations for the feature representation network. These augmentations encourage the encoder to learn node embeddings that capture hierarchical information shared across multi-resolution graphs.

\section{Multi-Resolution Graph Contrastive Learning}
\label{ch5:methods}
This section presents our approach to constructing diffusion filters on a graph. 

Graph tasks such as node classification rely on the assumption that nodes with the same labels have similar features. This assumption serves as an inductive bias for training an encoder that learns a lower-dimensional latent space, where similar nodes are mapped to nearby locations. However, the notion of similarity can vary across different graphs. In homophilic graphs, nodes in close proximity tend to have similar labels, whereas, in heterophilic graphs, the similarity is determined by the structural equivalence of the nodes. Our objective is to learn an encoder that captures various levels of structural similarity between pairs of nodes in a self-supervised manner.

Much of the existing methods attempt to capture structural information by relying on predefined notions of similarity, which may fall short when dealing with complex structural equivalence~\citep{yanardag2015deep, ribeiro2017struc2vec}. Diffusion methods quantify similarity at different time scales using the notion of diffusion distance. This metric measures the heat diffused between nodes $i$ and $j$ within a given time $t$ and vice versa. However, as $t$ increases, the diffusion process is dominated by low-frequency components, thereby neglecting higher-order structural information. Alternatively, diffusion wavelet filters~\citep{coifman2006diffusion} capture a band-pass response of a signal on a graph; that is, each filter reflects the diffusion distance in a time interval. The filters are constructed on a dyadic scale ordered as fine to coarse-grained. Thus explicitly highlighting the higher-order structural information encoded in a graph.

Our work leverages diffusion wavelet filters to construct augmentations of the graph and utilise it with a multi-view contrastive objective for learning resolution invariant representations. Our experimental results on structural equivalence tasks demonstrate that the learned encodings can capture a varying level of hierarchical relationship within graphs. Figure~\ref{struct_eqv} illustrates our approach using a synthetic graph, where nodes are coloured based on their structural roles, and the right side depicts the first two principal components of the learned representations. We observe that structurally equivalent nodes are grouped together in the representation space, showcasing the benefits of employing multi-resolution views. The overall pipeline of our approach is presented in Figure~\ref{fig:flowdiag}. We next introduce diffusion wavelets and provide details of the training setup employed in our paper.
\subsection{Multi-resolution graph augmentations}
Let us consider an undirected graph $\gG = (\gV, \gE)$ with $N = |\gV|$ nodes, $|\gE|$ edges, and $\XBRV \in \mathds{R}^{N\times d_i}$ as the node attribute matrix. Let $\ABRV \in \mathds{R}^{N\times N}$ be a weighted adjacency matrix, $\DBRV = \text{diag}(d_1, \ldots, d_N)$ as the degree matrix where each diagonal entry is the degree of the respective node, $\tilde{\ABRV}_1 = \ABRV \DBRV^{-1}$ as the column-normalised adjacency matrix, and $\TBRV = \alpha\IBRV_n + (1-\alpha)\tilde{\ABRV}_1$ as the normalised lazy random walk diffusion matrix with a restart probability $\alpha \in [0.5, 1]$.

We utilise the lazy diffusion operator to introduce multi-resolution filters on the graph, following the approach of~\citet{coifman2006diffusion} and~\citet{gama2018diffusion}. To put it formally, 
\begin{align}
\PhiBRV_1 = \IBRV - \TBRV ,\quad \PhiBRV_j = \TBRV^{2^{j-1}}(\IBRV - \TBRV^{2^{j-1}}), \quad j>0.
\end{align}
Here, $\PhiBRV_1, \ldots, \PhiBRV_{K-1}$ represents a bank of filters on a dyadic scale where each filter emphasises the higher-order connectivity information of a graph at different granularity. The restart probability $\alpha$ controls the influence of the local structure and the rate of diffusion propagation in the filters. This sequence of filters provides a multi-resolution view of the graph, which we later utilise for training the encoder neural network.

We also sparsify the filters using a threshold $\epsilon$ that sets the entries below $\epsilon$ to zero, resulting in $\ABRV_k = \PhiBRV_k [\PhiBRV_k<\epsilon]$, where $[\PhiBRV_k<\epsilon]$ represents element-wise comparison. We then normalise $\ABRV_k$ to obtain $\tilde{\ABRV}_k$. Both $\epsilon$ and the restart probability $\alpha$ are treated as hyperparameters.
\subsection{Feature representation network}
Given a set $\tilde{\ABRV} = \{\tilde{\ABRV}1, \cdots, \tilde{\ABRV}{K}\}$ of $K$ augmentation views of the graph. We combine each view with the attribute matrix and utilise an encoder network $f_{\omega}^k: \mathds{R}^{N\times d_i} \times \mathds{R}^{N\times N} \rightarrow \mathds{R}^{d}$ to map them to their respective node feature representation matrix $\HBRV_k \in \mathds{R}^{N\times d}$. The individual node representations from each view are then aggregated using a readout operation to obtain $K$ summary representations. These summary representations are further projected to obtain a graph level representation $\zBRV_k$ using a shared projection network $g_{\theta}: \mathds{R}^{N\times d} \rightarrow \mathds{R}^{d_z}$. Formally,
\begin{align}
\HBRV_k &= f_{\omega} (\XBRV, \tilde{\ABRV}k), \\
\vBRV_k &= \text{ReadOut}(\HBRV_k), \\
\zBRV_k &= g_{\theta} (\vBRV_k), \quad \forall k \geq 1.
\end{align}
The encoder $f_{\omega}^k$ is implemented as a two-layer GCN with parameters $\omega$ and a PReLU non-linearity~\citep{He_2015_ICCV}. The projection network $g_{\theta}$ is implemented as a two-layer MLP with parameters $\theta$, also utilising a PReLU non-linearity.
\subsection{Contrastive objective}
We train the encoder neural network using a local-global contrastive strategy, which involves contrasting the node representation of one view with the graph representation of other views, so, for $K$ different views, we have $\binom{K}{2}$ pairs. The negative samples for each view of the graph are generated using a stochastic corruption function from~\citet{velickovic2019deep} given as $(\XBRV^c,\tilde{\ABRV}^c) = C(\XBRV, \tilde{\ABRV})$, where $\HBRV_k^c$ represents the node-level representation of the corrupted nodes for the respective views.

The final optimisation objective is a binary cross-entropy loss with a noise contrastive objective, given by
\begin{align}
\gL &= \frac{1}{N+K} \sum_k \sum_{j, j\neq k} \Big( \sum_i \mathds{E}_{(\XBRV,\tilde{\ABRV}_k)} \left[\log D(\hBRV_i^j, \zBRV_k) \right] + \mathds{E}_{(\XBRV^c,\tilde{\ABRV}^c_j)} \left[\log (1 - D(\hBRV_i^{c}, \zBRV_k) \right] \Big).
\end{align}
where $D$ represents a \textit{discriminator} that maps a pair of a local node representation and a global graph representation to a real-valued score. This scoring mechanism acts as a proxy for mutual information and aims to assign a higher score to node-graph pairs from the same view and a lower score to pairs from different views. In line with existing works~\citep{velickovic2019deep, hassani2020contrastive}, we implement $D(\wBRV_i,\wBRV_j) = \langle\wBRV_i,\wBRV_j\rangle$ as a dot product.
\begin{table*}[t]
\setlength\tabcolsep{1.5pt}
\renewcommand{\arraystretch}{1.0}
    \centering
\begin{tabular}{l|lll|lll|lll}
      \toprule
         &\multicolumn{3}{c}{\textit{Proximity}} & \multicolumn{3}{c}{\textit{Structural}} & \multicolumn{3}{c}{\textit{Mixed}}\\
       & \bfseries Cora&\bfseries Citeseer & \bfseries PubMed & \bfseries Cornell & \bfseries Texas& \bfseries Wisconsin & \bfseries Squirrel & \bfseries Chameleon &\bfseries Actor\\
      \midrule
      Nodes & $3327$&$5429$ &$19717$ & $183$ &$183$  &$251$ &$5201$ & $2277$  & $7600$ \\
      Edges & $5429$&$4732$ &$44338$& $295$& $309$ & $499$ &$217073$ &$36101$ &$33544$ \\
      Attributes &$1433$&$3703$&$500$& $1703$  &$1703$ &$1703$&$2089$  & $2325$ & $931$ \\  
      Classes &$7$ &$6$ &$3$ &$5$ &$5$ &$5$ &$5$ &$5$&$5$ \\
      Homophily & $0.83$& $0.71$ & $0.79$ & $0.11$& $0.06$ & $0.16$& $0.22$& $0.25$ & $0.24$\\
      \bottomrule 
    \end{tabular}
    \caption{Summary statistics of different graphs used in our experiments. We also report the homophily score in the last row to outline the difference between various datasets. A higher value means graphs have a homophily property, and a lower value implies heterophily.}
    \label{tab:stats}
\end{table*}

\section{Experiments and Results}
\label{ch5:results}
\subsection{Implementation details and datasets} 
Our codebase is implemented in Python, utilising the torch geometric~\citep{Fey/Lenssen/2019} and PyGCL~\citep{Zhu:2021tu} (open-source library built on top of PyTorch~\citep{NEURIPS2019_9015}). All experiments were conducted on Nvidia GeForce RTX 2080 GPUs with $11$ GB memory. For training purpose following~\citep{velickovic2019deep, hassani2020contrastive, tang2022graph}, we use the Adam~\citep{kingma2014adam} optimiser with a learning rate of $0.001$. The restart probability $\alpha$ was set to $0.2$ across all experiments.

We want to remind the readers that our main objective is to leverage higher-order structural information for node representation learning, which is particularly important for heterophily graphs. To investigate our approach, we initially conducted experiments on \textbf{synthetic structural graphs} following the methodology of~\citet{donnat2018learning}. These synthetic datasets begin with simple equivalence structures such as \textit{House}, \textit{Fan}, and \textit{Star}, which are subsequently interconnected along a cycle of predefined length to form an entire graph. The nodes are labelled according to their structural roles, and the node degree is used as the node attribute. 

Next, we considered \textbf{real-world structural graphs} from the WebKB~\citep{ghani2001cmu} database. WebKB comprises datasets of webpage connectivity within computer science departments across various universities. The nodes represent web pages, and the edges denote hyperlinks connecting them. The labels correspond to five structural roles: student, project, course, staff, and faculty. The node attributes are derived from the bag-of-words representation of the webpage content. We utilised the Cornell, Texas, and Wisconsin graphs from the WebKB dataset for our experiments.

The multi-resolution views provide a lens to look at a graph on varying scales. By simply reducing the number of filters, we can encourage the encoder to represent the local information content without losing generality. This property makes our approach more universal and suitable for both proximal and structural graphs. Hence, we also conducted experiments on commonly used \textbf{proximal graphs}: Cora, Citeseer, and Pubmed citation networks~\citep{sen2008collective,narayanan2016subgraph2vec}, as well as \textbf{mixed graphs}: Chameleon and Squirrel from the Wikidatabase~\citep{rozemberczki2021multi}, and the Actor graph introduced in~\citet{donnat2018learning}. In the proximal graphs, the nodes represent publications, and the edges represent citations. The bag-of-words representation of the publication content gives the node attributes. In the mixed graph, the nodes correspond to Wikipedia pages, the edges represent links between pages, and the attributes are bag-of-words representations of the nouns on each page. In the actor graph, the nodes represent actors, and the edges are based on the co-occurrence of their web pages. The dataset statistics are summarised in Table~\ref{tab:stats}. Additionally, we report an edge homophily score, which estimates the level of proximal or structural information within each dataset. For a graph $\gG=(V, E)$, the score is defined as,
\begin{equation}
\textbf{Homophily} = \frac{\sum_{(v_i,v_j)\in E} \mathds{I}(v_i=v_j)}{|E|}.
\end{equation}
where $\mathds{I}(.)$ is an indicator function that returns $1$ if the condition is true, otherwise $0$. An overall score close to $1$ indicates a more proximal graph, while a score close to $0$ suggests a more structural graph.
\subsection{Baselines and evaluation setup} 
In this work, we compare our approach against various unsupervised representation learning methods, including, DeepWalk~\citep{perozzi2014deepwalk}, node2vec~\citep{grover2016node2vec}, RolX~\citep{henderson2012rolx}, struc2vec~\citep{ribeiro2017struc2vec}, GraphWave~\citep{donnat2018learning}, GAE and its probabilistic counterpart VGAE~\citep{kipf2016variational}, ARGVA~\citep{pan2018adversarially}, DGI~\citep{velickovic2019deep}, GraphCL~\citep{you2020graph}, MVGRL~\citep{hassani2020contrastive} and NWR-GAE~\citep{tang2022graph}. 

To ensure a fair comparison, similar to previous works~\citep{velickovic2019deep, hassani2020contrastive}, we set the node representation size to $512$ and employ the same architecture for our encoder network. For our multi-resolution views, we conduct two sets of experiments. In the first set, we utilise a dedicated GCN for each view, while in the second set, we employ a shared GCN across all views. We only use dedicated GCN on synthetic datasets, and in real-world experiments, we report results for both dedicated and shared GCN in the encoder network. 

We utilise clustering metrics similar to existing approaches~\citep{donnat2018learning, tang2022graph} to evaluate the performance on synthetic datasets. For real datasets, we assess the performance of downstream tasks of node classification. For classification purposes, we use a logistic regression model. To ensure consistency, we create $10$ random splits, with $60\%$ of the data allocated for training, $20\%$ for validation, and the remaining portion for testing. The performance measure is reported as the mean and standard deviation of accuracy. The results for all datasets are summarised in Table~\ref{ch5tab:quantitative}.
\begin{sidewaystable}
\setlength\tabcolsep{1.5pt}
\renewcommand{\arraystretch}{1.25}
    \small
    \centering
    \scalebox{0.85}{
    \begin{tabular}{l|c|llllllllllllll}
      \toprule
      \bfseries Dataset & \bfseries Metrics &\bfseries DeepWalk &  \bfseries node2vec & \bfseries RolX & \bfseries struc2vec & \bfseries GraphWave & \bfseries GAE & \bfseries VGAE & \bfseries ARGVA & \bfseries DGI & \bfseries GraphCL & \bfseries MVGRL & \bfseries NWR-GAE & \bfseries Ours\\
      \midrule
      \cmidrule{2-15}
      House & Homogeneity & $0.01$ & $0.01$ & $\mathbf{1.0}$ & $0.99$ & $\mathbf{1.0}$ & $\mathbf{1.0}$ & $0.25$ & $0.28$ & $\mathbf{1.0}$ & $\mathbf{1.0}$ & $\mathbf{1.0}$ & $\mathbf{1.0}$ & $\mathbf{1.0}$ \\
      & Completeness & $0.01$ & $0.01$ & $\mathbf{1.0}$ & $0.99$ & $\mathbf{1.0}$ & $\mathbf{1.0}$ & $0.27$ & $0.28$ & $\mathbf{1.0}$ & $\mathbf{1.0}$ & $\mathbf{1.0}$ & $\mathbf{1.0}$ & $\mathbf{1.0}$\\
    & Silhouette & $0.29$ & $0.33$ & $\mathbf{0.99}$ & $0.45$ & $\mathbf{0.99}$ & $\mathbf{0.99}$ & $0.21$ & $0.19$ & $\mathbf{0.99}$ & $\mathbf{0.99}$ & $\mathbf{0.99}$ & $\mathbf{0.99}$ & $\mathbf{0.99}$\\
    \midrule
  House Perturbed & Homogeneity & $0.06$ & $0.03$ & $0.65$ & $0.21$ & $0.54$ & $0.36$ & $0.29$ & $0.24$ & $0.24$ & $0.41$ & $0.68$ & $0.60$ & $\mathbf{0.78}$\\
      & Completeness & $0.06$ & $0.03$ & $0.69$ & $0.24$ & $0.56$ & $0.37$ & $0.29$ & $0.24$ & $0.25$ & $0.42$ & $0.69$ & $0.62$ & $\mathbf{0.79}$\\
    & Silhouette & $0.25$ & $0.28$ & $0.51$ & $0.18$ & $0.37$ & $0.69$ & $0.21$ & $0.19$ & $0.44$ & $0.43$ & $0.57$ & $0.49$ & $\mathbf{0.76}$ \\
      \midrule 
  Varied & Homogeneity & $0.26$ & $0.23$ & $0.91$ & $0.63$ & $0.83$ & $0.65$ & $0.50$ & $0.66$ & $0.36$ & $0.93$ & $\mathbf{0.93}$ & $\mathbf{0.93}$ & $\mathbf{0.93}$\\
      & Completeness & $0.23$ & $0.22$ & $0.93$ & $0.58$ & $0.85$ & $0.69$ & $0.36$ & $0.57$ & $0.37$ & $0.89$ & $0.89$ & $\mathbf{0.94}$ &  $\mathbf{0.94}$\\
    & Silhouette & $0.35$ & $0.40$ & $0.82$ & $0.24$ & $0.81$ & $0.83$ & $0.21$ & $0.23$ & $0.94$ & $0.93$ & $0.90$ & $0.95$ & $\mathbf{0.97}$\\ 
    \midrule
 Varied Perturbed & Homogeneity & $0.30$ & $0.30$ & $0.74$ & $0.46$ & $0.69$ & $0.44$ & $0.42$ & $0.57$ & $0.36$ & $0.70$ & $0.73$ & $0.78$ & $\mathbf{0.83}$\\
  & Completeness & $0.27$ & $0.27$ & $0.72$ & $0.43$ & $0.68$ & $0.45$ & $0.43$ & $0.49$ & $0.36$ & $0.63$ & $0.67$ & $0.81$ & $\mathbf{0.86}$\\
    & Silhouette & $0.33$ & $0.36$ & $0.61$ & $0.29$ & $0.51$ & $0.52$ & $0.21$ & $0.20$ & $0.45$ & $0.69$ & $0.61$ & $0.84$ & $\mathbf{0.89}$\\    
    \bottomrule
    \end{tabular}}
    \caption{Here, we compare our approach with other baselines on the task of structural role identification for synthetic dataset introduced in~\citep{donnat2018learning}. Our approach consistently outperforms the baselines with three filter views.}
    \label{tab:toydata}
\end{sidewaystable}
\subsection{Results and discussion}
\subsubsection{Synthetic graphs} 
For this particular task, we adopt the clustering procedure outlined by~\citet{donnat2018learning}, and \citet{tang2022graph}. We employ agglomerative clustering with a single linkage to cluster the node representations and report three evaluation metrics: \textit{Homogeneity} (which measures the conditional entropy of the ground-truth labels given the predicted clusters), \textit{Completeness} (which calculates the ratio of samples with the same ground-truth label assigned to the same group), and \textit{Silhouette score} (which compares the intra-cluster distance to the inter-cluster distance).

In this set of experiments, we set the number of filters to three ($K=3$). The results for all synthetic graphs are presented in detail in Table~\ref{tab:toydata}. On the \textit{House} graph, our approach yields comparable performance to RolX, GraphWave, GAE, DG, GraphCL, MVGRL, and NWR-GAE. This outcome can be attributed to the simplicity of the graphs, resulting in high scores across most methods. However, the performance of the baseline methods drops significantly on other tasks, except for NWR-GAE. In a majority of cases, our approach outperforms NWR-GAE. On the \textit{Varied} graph, we achieve comparable Homogeneity and Completeness scores while outperforming the Silhouette score, indicating denser clusters in our case. These results provide compelling evidence that our approach is the best suitable for structural tasks.

We observed that increasing the number of views did not lead to improvement, which could be due to the toy nature of the datasets.
\begin{sidewaystable}
\setlength\tabcolsep{1.5pt}
\renewcommand{\arraystretch}{1.25}
\centering
    \scalebox{0.91}{
    \begin{tabular}{l|ccc|ccc|ccc}
      \toprule
      \bfseries Method &\multicolumn{3}{c}{\textit{Proximity}} & \multicolumn{3}{c}{\textit{Structural}} & \multicolumn{3}{c}{\textit{Mixed}}\\
      &\bfseries Cora& \bfseries Citeseer & \bfseries PubMed& \bfseries Cornell & \bfseries Texas & \bfseries Wisconsin & \bfseries Chameleon & \bfseries Squirrel & \bfseries Actor\\
      \midrule
    DeepWalk~\citep{perozzi2014deepwalk}  & $82.97\pm 1.67$& $68.99\pm 0.95$& $82.39\pm 4.88$ &$41.21\pm 3.40$ & $41.89\pm 7.81$& $43.62\pm 2.46$ & $68.03\pm 2.13$& $59.22\pm 2.35$ & $23.84\pm 2.14$\\
node2vec~\citep{grover2016node2vec} & $81.93\pm 1.43$& $64.56\pm 1.65$& $81.02\pm 1.48$ &$40.54\pm 1.62$ & $48.64\pm 2.92$& $36.27\pm 2.08$ & $65.67\pm 2.31$ & $48.29\pm 1.67$& $24.14\pm 1.02$\\
    RolX~\citep{henderson2012rolx} & $29.70\pm 2.89$& $20.90\pm 0.72$& $39.85\pm 2.33$ &$25.67\pm 11.78$ & $42.56\pm 7.13$& $24.92\pm 13.43$ & $22.75\pm 2.12$ & $20.50\pm 1.18$& $25.42\pm 0.55$\\
    struc2vec~\citep{ribeiro2017struc2vec} & $41.46\pm 1.49$& $51.70\pm 0.67$& $81.49\pm 0.33$ &$23.72\pm 13.69$ & $47.29\pm 7.21$& $24.59\pm 12.14$ & $60.63\pm 2.90$& $52.59\pm 0.69$& $25.13\pm 0.79$\\
    GraphWave~\citep{donnat2018learning}  & $28.83\pm 2.39$& $20.79\pm 1.59$& $20.96\pm 2.35$ &$45.96\pm 2.20$ & $37.45\pm 7.09$& $39.24\pm 5.16$ & $17.59\pm 3.42$ & $25.69\pm 0.53$ & $27.29\pm 3.09$\\
    GAE~\citep{kipf2016variational} & $72.06\pm 2.54$& $57.10\pm 1.62$& $73.24\pm 0.88$ &$45.40\pm 9.99$ & $58.78\pm 3.41$& $34.11\pm 8.06$ & $22.03\pm 1.09$ & $29.34\pm 1.12$ & $28.63\pm 1.05$\\  
    VGAE~\citep{kipf2016variational} & $72.87 \pm 1.48$& $60.78 \pm 1.92$& $81.34 \pm 0.79$ &$49.32 \pm 9.19$ & $39.18 \pm 8.96$& $38.27 \pm 6.12$ &$ 20.17 \pm 1.30$& $19.57 \pm 1.63$ &  $26.41 \pm 1.07$\\
    ARGVA~\citep{pan2018adversarially}& $72.88 \pm 3.83$& $63.36 \pm 2.08$& $75.32 \pm 0.63$ &$ 41.08 \pm 4.85$ & $43.24 \pm 5.38 $& $41.17 \pm 5.20 $ & $21.17 \pm 0.78$&$20.61 \pm 0.73$  & $28.97 \pm 1.17$\\
      DGI~\citep{velickovic2019deep} & $ 84.76 \pm 1.39$& $ 71.68 \pm 1.54$& $84.29 \pm 1.07$ &$46.48 \pm 7.97$ & $52.97 \pm 5.64$& $55.68 \pm 2.97$ & $25.89 \pm 1.49$ & $25.89 \pm 1.62$ & $ 20.45 \pm 1.32$ \\
    GraphCL & $84.23 \pm 1.51$& $ 73.51 \pm 1.73 $& $82.59 \pm 0.71$ &$ 44.86 \pm 3.73$ & $46.48 \pm 5.85$& $53.72 \pm 1.07 $ & $26.27 \pm 1.53$&$21.32 \pm 1.66$&$28.64 \pm 1.28 $\\   
      MVGRL & $ 86.23 \pm 2.71 $&$ 73.81 \pm 1.53$& $83.94 \pm 0.75$& $ 53.51 \pm 3.26$ & $56.75 \pm 5.97$ & $ 57.25 \pm 5.94$ & $58.73 \pm 2.03$& $40.64 \pm 1.15 $ & $ 31.07 \pm 0.29$\\
     NWR-GAE &$ 83.62 \pm 1.61 $ & $ 71.45 \pm 2.41 $ &$83.44 \pm 0.92$&$58.64 \pm 5.61$ &$69.62 \pm 6.66$ & $68.23 \pm 6.11$ &$\mathbf{ 72.04 \pm 2.59} $ & $\mathbf{64.81 \pm 1.83}$ & $30.17 \pm 0.17$\\
     \midrule
      Ours (dedicated) &$\mathbf{87.67\pm 1.49}$ & $\mathbf{75.41\pm 1.11}$ &$87.3\pm 0.30$&$66.84\pm 7.01$ &$\mathbf{72.89\pm 6.60}$ & $\mathbf{76.27\pm 4.06}$ &$64.54\pm 2.25$ & $43.17\pm 1.11$ & $\mathbf{36.01\pm 0.68}$\\  
      Ours (shared) &$85.96\pm 1.75$ & $75.22\pm 0.99$ &$\mathbf{87.65\pm 0.33}$&$\mathbf{74.70\pm 7.56}$ &$72.36\pm 3.76$ & $72.35\pm 4.75$ &$61.49\pm 2.14$ & $42.87\pm 1.03$ & $35.60\pm 0.80$\\         
      \bottomrule 
    \end{tabular}}
    \caption{We compare the performance of our approach with various baselines on node classification. The first task is the proximal graphs -- nodes in the local neighbourhood share the same label. Next are the structural graphs -- nodes with structural similarity share the same label, and lastly, mixed graphs with both types of node labels. Our setup works best across all three graph categories and significantly outperforms baselines on structural graphs. We note that the performance of NWR-GAE is best on the squirrel.}
    \label{ch5tab:quantitative}
\end{sidewaystable}

\subsubsection{Structural, Mixed and Proximal graphs}
The comparison of node classification accuracy with baseline methods across three types of graph data is presented in Table~\ref{ch5tab:quantitative}. We utilise (K=$4$) multi-resolution views for structural and mixed graphs, while for proximal graphs, we employ (K=$2$) views in addition to the local graph adjacency. Interestingly, we observed that the performance did not improve when including more filters.

On structural graphs, our approach outperforms the baselines significantly, achieving an improvement of $16.06\%$ on Cornell (using shared GCN), $3.27\%$ on Texas (using dedicated GCN), and $8.04\%$ on Wisconsin compared to the best baseline method NWR-GAE. Furthermore, our approach surpasses all baselines on proximal graphs, with a $3.7\%$ improvement on the PubMed graph, which is notably significant in size. These results demonstrate the universality of our approach.

On mixed data, our approach outperforms NWR-GAE by $5.84\%$ on the actor dataset. However, NWR-GAE achieves better performance on the chameleon and squirrel datasets. We hypothesise that this can be attributed to NWR-GAE using a more expressive decoder and additional loss terms based on degree and neighbourhood size.

Regarding our two encoder variants, we observe that the GCN \textit{shared} variant achieves higher performance on the Cornell and PubMed tasks. Additionally, it performs competitively on other tasks. This result can be attributed to the explicit frequency resolution provided by the multi-resolution views, which helps prevent representation smoothing often observed with GCN networks~\citep{chen2020measuring}. On the other hand, the GCN \textit{dedicated} variant, which employs a separate encoder for each view, emerges as the overall best method.

The importance of representations is primarily determined by their applications to downstream tasks. By allowing the feature representation network to learn the appropriate scale invariance, we can tailor the number of views accordingly. For example, we can reduce the number of filter views in graphs where local connectivity is more crucial. Conversely, introducing more filters can improve performance in graphs where structural information is more relevant. Overall, the multi-resolution views are a simple and efficient strategy for training self-supervised representation learners.

Next, we investigate the effect of including higher-order resolution views on the performance of structural datasets.

\subsubsection{Ablation Study}
\label{sec:ablation}
\begin{figure}
    \centering
    \includegraphics[width=0.32\columnwidth]{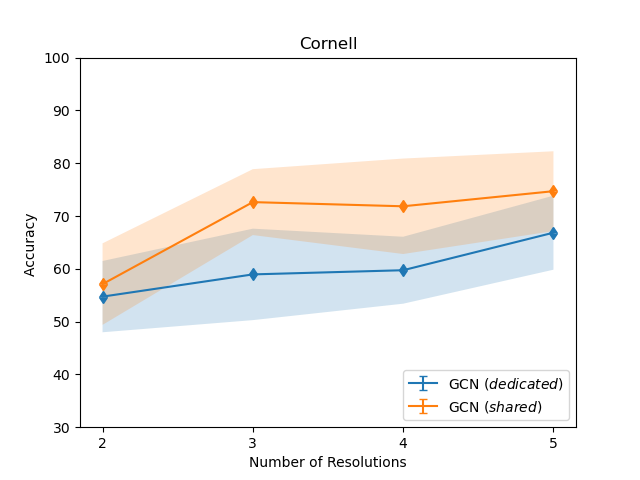}
    \includegraphics[width=0.32\columnwidth]{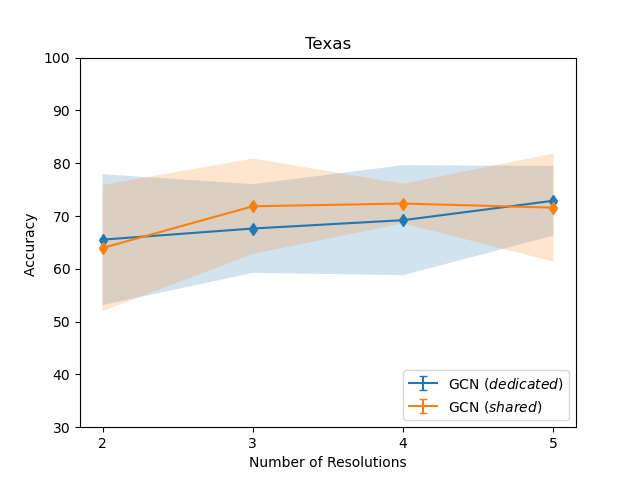}
    \includegraphics[width=0.32\columnwidth]{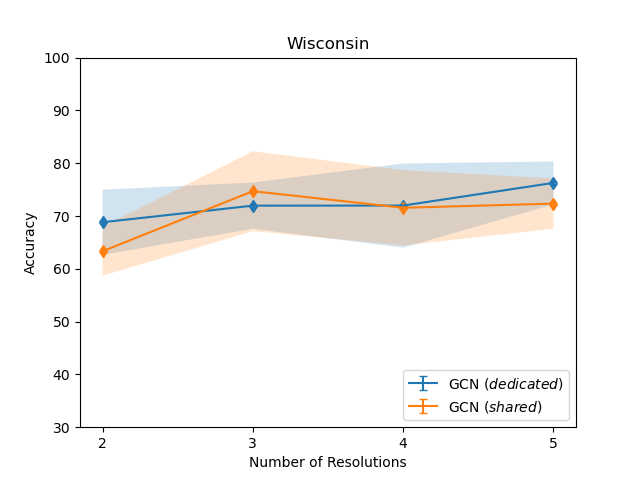}
    \caption{Ablation Study. Here, we investigate the effect of including higher-resolution graph views on top of the local graph structure. We report node classification accuracy averaged across ten random splits on structure graphs with an increasing number of views.}
    \label{fig:ablation}
\end{figure}

We conducted ablation experiments to examine the impact of increasing the number of resolutions on the downstream performance. We only use structural data for this purpose since we expect higher-order filters to be more beneficial for learning structural information. The results on structural data are illustrated in Figure~\ref{fig:ablation}. 

We observed that increasing the resolution leads to improved performance. However, beyond a certain level, the performance gain becomes marginal. We hypothesise that this phenomenon arises because the same node attributes are used for message passing across different scales. Due to a dyadic scale, the coarse-grained filters get sparse, and the message passing may smooth out and not update further. We wish to investigate this phenomenon in future work.

\section{Conclusion}
\label{ch5:futurework}
This paper introduced wavelet diffusion filters as a means to incorporate multi-resolution graph views in self-supervised learning for heterophilic graphs. These filters have desirable properties, including stability to signal deformation and perturbation, as well as the ability to capture higher-order structural regularities within graphs. Utilising these filters as graph augmentations in the contrasting training objective introduces a
resolution invariance in the learned representations, which proves advantageous for addressing structural equivalence problems. Through empirical experiments on both synthetic and real-world structural datasets, we demonstrated the effectiveness of our approach. Importantly, our method also outperformed baseline approaches on proximal graphs, highlighting its general applicability for learning representations of data on a broad range of graphs. 
\newpage
\bibliography{references}

\begin{thebibliography}{}

\bibitem[Al-Rfou et~al., 2019]{al2019ddgk}
Al-Rfou, R., Perozzi, B., and Zelle, D. (2019).
\newblock Ddgk: Learning graph representations for deep divergence graph
  kernels.
\newblock In {\em The World Wide Web Conference}, pages 37--48.

\bibitem[Belkin and Niyogi, 2003]{belkin2003laplacian}
Belkin, M. and Niyogi, P. (2003).
\newblock Laplacian eigenmaps for dimensionality reduction and data
  representation.
\newblock {\em Neural computation}, 15(6):1373--1396.

\bibitem[Belkin et~al., 2006]{belkin2006manifold}
Belkin, M., Niyogi, P., and Sindhwani, V. (2006).
\newblock Manifold regularization: A geometric framework for learning from
  labeled and unlabeled examples.
\newblock {\em Journal of machine learning research}, 7(11).

\bibitem[Bojchevski and G{\"u}nnemann, 2017]{bojchevski2017deep}
Bojchevski, A. and G{\"u}nnemann, S. (2017).
\newblock Deep gaussian embedding of graphs: Unsupervised inductive learning
  via ranking.
\newblock {\em arXiv preprint arXiv:1707.03815}.

\bibitem[Borgwardt et~al., 2020]{borgwardt2020graph}
Borgwardt, K., Ghisu, E., Llinares-L{\'o}pez, F., O'Bray, L., and Rieck, B.
  (2020).
\newblock Graph kernels: State-of-the-art and future challenges.
\newblock {\em arXiv preprint arXiv:2011.03854}.

\bibitem[Bui et~al., 2018]{bui2018neural}
Bui, T.~D., Ravi, S., and Ramavajjala, V. (2018).
\newblock Neural graph learning: Training neural networks using graphs.
\newblock In {\em Proceedings of the Eleventh ACM International Conference on
  Web Search and Data Mining}, pages 64--71.

\bibitem[Cai et~al., 2018]{cai2018comprehensive}
Cai, H., Zheng, V.~W., and Chang, K. C.-C. (2018).
\newblock A comprehensive survey of graph embedding: Problems, techniques, and
  applications.
\newblock {\em IEEE Transactions on Knowledge and Data Engineering},
  30(9):1616--1637.

\bibitem[Cangea et~al., 2018]{cangea2018towards}
Cangea, C., Veli{\v{c}}kovi{\'c}, P., Jovanovi{\'c}, N., Kipf, T., and Li{\`o},
  P. (2018).
\newblock Towards sparse hierarchical graph classifiers.
\newblock {\em arXiv preprint arXiv:1811.01287}.

\bibitem[Chen et~al., 2020a]{chen2020measuring}
Chen, D., Lin, Y., Li, W., Li, P., Zhou, J., and Sun, X. (2020a).
\newblock Measuring and relieving the over-smoothing problem for graph neural
  networks from the topological view.
\newblock In {\em Proceedings of the AAAI Conference on Artificial
  Intelligence}, volume~34, pages 3438--3445.

\bibitem[Chen et~al., 2020b]{chen2020graph}
Chen, F., Wang, Y.-C., Wang, B., and Kuo, C.-C.~J. (2020b).
\newblock Graph representation learning: a survey.
\newblock {\em APSIPA Transactions on Signal and Information Processing}, 9.

\bibitem[Chen et~al., 2020c]{chen2020simple}
Chen, T., Kornblith, S., Norouzi, M., and Hinton, G. (2020c).
\newblock A simple framework for contrastive learning of visual
  representations.
\newblock In {\em International conference on machine learning}, pages
  1597--1607. PMLR.

\bibitem[Chen et~al., 2017]{chen2017sampling}
Chen, T., Sun, Y., Shi, Y., and Hong, L. (2017).
\newblock On sampling strategies for neural network-based collaborative
  filtering.
\newblock In {\em Proceedings of the 23rd ACM SIGKDD International Conference
  on Knowledge Discovery and Data Mining}, pages 767--776.

\bibitem[Chuang et~al., 2020]{chuang2020debiased}
Chuang, C.-Y., Robinson, J., Lin, Y.-C., Torralba, A., and Jegelka, S. (2020).
\newblock Debiased contrastive learning.
\newblock {\em Advances in neural information processing systems},
  33:8765--8775.

\bibitem[Coifman and Maggioni, 2006]{coifman2006diffusion}
Coifman, R.~R. and Maggioni, M. (2006).
\newblock Diffusion wavelets.
\newblock {\em Applied and computational harmonic analysis}, 21(1):53--94.

\bibitem[Dai et~al., 2016]{dai2016discriminative}
Dai, H., Dai, B., and Song, L. (2016).
\newblock Discriminative embeddings of latent variable models for structured
  data.
\newblock In {\em International conference on machine learning}, pages
  2702--2711. PMLR.

\bibitem[Defferrard et~al., 2016]{defferrard2016convolutional}
Defferrard, M., Bresson, X., and Vandergheynst, P. (2016).
\newblock Convolutional neural networks on graphs with fast localized spectral
  filtering.
\newblock {\em Advances in neural information processing systems}, 29.

\bibitem[Donnat et~al., 2018]{donnat2018learning}
Donnat, C., Zitnik, M., Hallac, D., and Leskovec, J. (2018).
\newblock Learning structural node embeddings via diffusion wavelets.
\newblock In {\em Proceedings of the 24th ACM SIGKDD International Conference
  on Knowledge Discovery \& Data Mining}, pages 1320--1329.

\bibitem[Fey and Lenssen, 2019]{Fey/Lenssen/2019}
Fey, M. and Lenssen, J.~E. (2019).
\newblock Fast graph representation learning with {PyTorch Geometric}.
\newblock In {\em ICLR Workshop on Representation Learning on Graphs and
  Manifolds}.

\bibitem[Gama et~al., 2018]{gama2018diffusion}
Gama, F., Ribeiro, A., and Bruna, J. (2018).
\newblock Diffusion scattering transforms on graphs.
\newblock {\em arXiv preprint arXiv:1806.08829}.

\bibitem[Gao et~al., 2019]{gao2019geometric}
Gao, F., Wolf, G., and Hirn, M. (2019).
\newblock Geometric scattering for graph data analysis.
\newblock In {\em International Conference on Machine Learning}, pages
  2122--2131. PMLR.

\bibitem[Ghani, 2001]{ghani2001cmu}
Ghani, R. (2001).
\newblock Cmu world wide knowledge base (webkb) project.
\newblock {\em Online Referencing: http://www. cs. cmu. edu/\~{} webkb/(Access
  Date: 22 June 2018)}.

\bibitem[Grover and Leskovec, 2016]{grover2016node2vec}
Grover, A. and Leskovec, J. (2016).
\newblock node2vec: Scalable feature learning for networks.
\newblock In {\em Proceedings of the 22nd ACM SIGKDD international conference
  on Knowledge discovery and data mining}, pages 855--864.

\bibitem[Grover et~al., 2019]{grover2019graphite}
Grover, A., Zweig, A., and Ermon, S. (2019).
\newblock Graphite: Iterative generative modeling of graphs.
\newblock In {\em International conference on machine learning}, pages
  2434--2444. PMLR.

\bibitem[Hamilton et~al., 2017]{hamilton2017inductive}
Hamilton, W., Ying, Z., and Leskovec, J. (2017).
\newblock Inductive representation learning on large graphs.
\newblock {\em Advances in neural information processing systems}, 30.

\bibitem[Hamilton, 2020]{hamilton2020graph}
Hamilton, W.~L. (2020).
\newblock Graph representation learning.
\newblock {\em Synthesis Lectures on Artifical Intelligence and Machine
  Learning}, 14(3):1--159.

\bibitem[Hassani and Khasahmadi, 2020]{hassani2020contrastive}
Hassani, K. and Khasahmadi, A.~H. (2020).
\newblock Contrastive multi-view representation learning on graphs.
\newblock In {\em International Conference on Machine Learning}, pages
  4116--4126. PMLR.

\bibitem[He et~al., 2020]{he2020momentum}
He, K., Fan, H., Wu, Y., Xie, S., and Girshick, R. (2020).
\newblock Momentum contrast for unsupervised visual representation learning.
\newblock In {\em Proceedings of the IEEE/CVF conference on computer vision and
  pattern recognition}, pages 9729--9738.

\bibitem[Henderson et~al., 2012]{henderson2012rolx}
Henderson, K., Gallagher, B., Eliassi-Rad, T., Tong, H., Basu, S., Akoglu, L.,
  Koutra, D., Faloutsos, C., and Li, L. (2012).
\newblock Rolx: structural role extraction \& mining in large graphs.
\newblock In {\em Proceedings of the 18th ACM SIGKDD international conference
  on Knowledge discovery and data mining}, pages 1231--1239.

\bibitem[Jaiswal et~al., 2020]{jaiswal2020survey}
Jaiswal, A., Babu, A.~R., Zadeh, M.~Z., Banerjee, D., and Makedon, F. (2020).
\newblock A survey on contrastive self-supervised learning.
\newblock {\em Technologies}, 9(1):2.

\bibitem[Kingma and Ba, 2014]{kingma2014adam}
Kingma, D.~P. and Ba, J. (2014).
\newblock Adam: A method for stochastic optimization.
\newblock {\em arXiv preprint arXiv:1412.6980}.

\bibitem[Kingma and Welling, 2013]{kingma2013auto}
Kingma, D.~P. and Welling, M. (2013).
\newblock Auto-encoding variational bayes.
\newblock {\em arXiv preprint arXiv:1312.6114}.

\bibitem[Kipf and Welling, 2016a]{kipf2016semi}
Kipf, T.~N. and Welling, M. (2016a).
\newblock Semi-supervised classification with graph convolutional networks.
\newblock {\em arXiv preprint arXiv:1609.02907}.

\bibitem[Kipf and Welling, 2016b]{kipf2016variational}
Kipf, T.~N. and Welling, M. (2016b).
\newblock Variational graph auto-encoders.
\newblock {\em arXiv preprint arXiv:1611.07308}.

\bibitem[Laskin et~al., 2020]{laskin2020curl}
Laskin, M., Srinivas, A., and Abbeel, P. (2020).
\newblock Curl: Contrastive unsupervised representations for reinforcement
  learning.
\newblock In {\em International Conference on Machine Learning}, pages
  5639--5650. PMLR.

\bibitem[Levie et~al., 2018]{levie2018cayleynets}
Levie, R., Monti, F., Bresson, X., and Bronstein, M.~M. (2018).
\newblock Cayleynets: Graph convolutional neural networks with complex rational
  spectral filters.
\newblock {\em IEEE Transactions on Signal Processing}, 67(1):97--109.

\bibitem[Li et~al., 2015]{li2015gated}
Li, Y., Tarlow, D., Brockschmidt, M., and Zemel, R. (2015).
\newblock Gated graph sequence neural networks.
\newblock {\em arXiv preprint arXiv:1511.05493}.

\bibitem[Lin et~al., 2020]{lin2020kgnn}
Lin, X., Quan, Z., Wang, Z.-J., Ma, T., and Zeng, X. (2020).
\newblock Kgnn: Knowledge graph neural network for drug-drug interaction
  prediction.
\newblock In {\em IJCAI}, volume 380, pages 2739--2745.

\bibitem[Mallat, 2012]{mallat2012group}
Mallat, S. (2012).
\newblock Group invariant scattering.
\newblock {\em Communications on Pure and Applied Mathematics},
  65(10):1331--1398.

\bibitem[Mikolov et~al., 2013]{mikolov2013distributed}
Mikolov, T., Sutskever, I., Chen, K., Corrado, G.~S., and Dean, J. (2013).
\newblock Distributed representations of words and phrases and their
  compositionality.
\newblock {\em Advances in neural information processing systems}, 26.

\bibitem[Narayanan et~al., 2016]{narayanan2016subgraph2vec}
Narayanan, A., Chandramohan, M., Chen, L., Liu, Y., and Saminathan, S. (2016).
\newblock subgraph2vec: Learning distributed representations of rooted
  sub-graphs from large graphs.
\newblock {\em arXiv preprint arXiv:1606.08928}.

\bibitem[Pan et~al., 2018]{pan2018adversarially}
Pan, S., Hu, R., Long, G., Jiang, J., Yao, L., and Zhang, C. (2018).
\newblock Adversarially regularized graph autoencoder for graph embedding.
\newblock {\em arXiv preprint arXiv:1802.04407}.

\bibitem[Park et~al., 2019]{park2019symmetric}
Park, J., Lee, M., Chang, H.~J., Lee, K., and Choi, J.~Y. (2019).
\newblock Symmetric graph convolutional autoencoder for unsupervised graph
  representation learning.
\newblock In {\em Proceedings of the IEEE/CVF International Conference on
  Computer Vision}, pages 6519--6528.

\bibitem[Paszke et~al., 2019]{NEURIPS2019_9015}
Paszke, A., Gross, S., Massa, F., Lerer, A., Bradbury, J., Chanan, G., Killeen,
  T., Lin, Z., Gimelshein, N., Antiga, L., Desmaison, A., Kopf, A., Yang, E.,
  DeVito, Z., Raison, M., Tejani, A., Chilamkurthy, S., Steiner, B., Fang, L.,
  Bai, J., and Chintala, S. (2019).
\newblock Pytorch: An imperative style, high-performance deep learning library.
\newblock In Wallach, H., Larochelle, H., Beygelzimer, A., d\textquotesingle
  Alch\'{e}-Buc, F., Fox, E., and Garnett, R., editors, {\em Advances in Neural
  Information Processing Systems 32}, pages 8024--8035. Curran Associates, Inc.

\bibitem[Paulus et~al., 2020]{NEURIPS2020_3df80af5}
Paulus, M., Choi, D., Tarlow, D., Krause, A., and Maddison, C.~J. (2020).
\newblock {Gradient Estimation with Stochastic Softmax Tricks}.
\newblock In Larochelle, H., Ranzato, M., Hadsell, R., Balcan, M., and Lin, H.,
  editors, {\em Advances in Neural Information Processing Systems}, volume~33,
  pages 5691--5704. Curran Associates, Inc.

\bibitem[Perozzi et~al., 2014]{perozzi2014deepwalk}
Perozzi, B., Al-Rfou, R., and Skiena, S. (2014).
\newblock Deepwalk: Online learning of social representations.
\newblock In {\em Proceedings of the 20th ACM SIGKDD international conference
  on Knowledge discovery and data mining}, pages 701--710.

\bibitem[Ribeiro et~al., 2017]{ribeiro2017struc2vec}
Ribeiro, L.~F., Saverese, P.~H., and Figueiredo, D.~R. (2017).
\newblock struc2vec: Learning node representations from structural identity.
\newblock In {\em Proceedings of the 23rd ACM SIGKDD international conference
  on knowledge discovery and data mining}, pages 385--394.

\bibitem[Rozemberczki et~al., 2021]{rozemberczki2021multi}
Rozemberczki, B., Allen, C., and Sarkar, R. (2021).
\newblock Multi-scale attributed node embedding.
\newblock {\em Journal of Complex Networks}, 9(2):cnab014.

\bibitem[Sen et~al., 2008]{sen2008collective}
Sen, P., Namata, G., Bilgic, M., Getoor, L., Galligher, B., and Eliassi-Rad, T.
  (2008).
\newblock Collective classification in network data.
\newblock {\em AI magazine}, 29(3):93--93.

\bibitem[Shervashidze et~al., 2011]{shervashidze2011weisfeiler}
Shervashidze, N., Schweitzer, P., Van~Leeuwen, E.~J., Mehlhorn, K., and
  Borgwardt, K.~M. (2011).
\newblock Weisfeiler-lehman graph kernels.
\newblock {\em Journal of Machine Learning Research}, 12(9).

\bibitem[Shi et~al., 2020]{shi2020effective}
Shi, H., Fan, H., and Kwok, J.~T. (2020).
\newblock Effective decoding in graph auto-encoder using triadic closure.
\newblock In {\em Proceedings of the AAAI Conference on Artificial
  Intelligence}, volume~34, pages 906--913.

\bibitem[Sun et~al., 2019]{sun2019infograph}
Sun, F.-Y., Hoffmann, J., Verma, V., and Tang, J. (2019).
\newblock Infograph: Unsupervised and semi-supervised graph-level
  representation learning via mutual information maximization.
\newblock {\em arXiv preprint arXiv:1908.01000}.

\bibitem[Tang et~al., 2022]{tang2022graph}
Tang, M., Yang, C., and Li, P. (2022).
\newblock Graph auto-encoder via neighborhood wasserstein reconstruction.
\newblock {\em arXiv preprint arXiv:2202.09025}.

\bibitem[Tong et~al., 2020]{tong2020data}
Tong, A., Wenkel, F., MacDonald, K., Krishnaswamy, S., and Wolf, G. (2020).
\newblock Data-driven learning of geometric scattering networks.
\newblock {\em arXiv preprint arXiv:2010.02415}.

\bibitem[Tsitsulin et~al., 2018]{tsitsulin2018verse}
Tsitsulin, A., Mottin, D., Karras, P., and M{\"u}ller, E. (2018).
\newblock Verse: Versatile graph embeddings from similarity measures.
\newblock In {\em Proceedings of the 2018 world wide web conference}, pages
  539--548.

\bibitem[Velickovic et~al., 2019]{velickovic2019deep}
Velickovic, P., Fedus, W., Hamilton, W.~L., Li{\`o}, P., Bengio, Y., and Hjelm,
  R.~D. (2019).
\newblock Deep graph infomax.
\newblock {\em ICLR (Poster)}, 2(3):4.

\bibitem[Wang et~al., 2017a]{wang2017mgae}
Wang, C., Pan, S., Long, G., Zhu, X., and Jiang, J. (2017a).
\newblock Mgae: Marginalized graph autoencoder for graph clustering.
\newblock In {\em Proceedings of the 2017 ACM on Conference on Information and
  Knowledge Management}, pages 889--898.

\bibitem[Wang et~al., 2022]{wang2022molecular}
Wang, Y., Wang, J., Cao, Z., and Barati~Farimani, A. (2022).
\newblock Molecular contrastive learning of representations via graph neural
  networks.
\newblock {\em Nature Machine Intelligence}, 4(3):279--287.

\bibitem[Wang et~al., 2021]{wang2021molclr}
Wang, Y., Wang, J., Cao, Z., and Farimani, A.~B. (2021).
\newblock Molclr: molecular contrastive learning of representations via graph
  neural networks.
\newblock {\em arXiv preprint arXiv:2102.10056}.

\bibitem[Wang et~al., 2017b]{wang2017predictive}
Wang, Z., Chen, C., and Li, W. (2017b).
\newblock Predictive network representation learning for link prediction.
\newblock In {\em Proceedings of the 40th international ACM SIGIR conference on
  research and development in information retrieval}, pages 969--972.

\bibitem[Weston et~al., 2012]{weston2012deep}
Weston, J., Ratle, F., Mobahi, H., and Collobert, R. (2012).
\newblock Deep learning via semi-supervised embedding.
\newblock In {\em Neural networks: Tricks of the trade}, pages 639--655.
  Springer.

\bibitem[Wu et~al., 2018]{wu2018moleculenet}
Wu, Z., Ramsundar, B., Feinberg, E.~N., Gomes, J., Geniesse, C., Pappu, A.~S.,
  Leswing, K., and Pande, V. (2018).
\newblock Moleculenet: a benchmark for molecular machine learning.
\newblock {\em Chemical science}, 9(2):513--530.

\bibitem[Yanardag and Vishwanathan, 2015]{yanardag2015deep}
Yanardag, P. and Vishwanathan, S. (2015).
\newblock Deep graph kernels.
\newblock In {\em Proceedings of the 21th ACM SIGKDD international conference
  on knowledge discovery and data mining}, pages 1365--1374.

\bibitem[Yang et~al., 2020]{yang2020secure}
Yang, C., Wang, H., Zhang, K., Chen, L., and Sun, L. (2020).
\newblock Secure deep graph generation with link differential privacy.
\newblock {\em arXiv preprint arXiv:2005.00455}.

\bibitem[You et~al., 2020]{you2020graph}
You, Y., Chen, T., Sui, Y., Chen, T., Wang, Z., and Shen, Y. (2020).
\newblock Graph contrastive learning with augmentations.
\newblock {\em Advances in Neural Information Processing Systems},
  33:5812--5823.

\bibitem[Zhang and Chen, 2018]{zhang2018link}
Zhang, M. and Chen, Y. (2018).
\newblock Link prediction based on graph neural networks.
\newblock {\em Advances in neural information processing systems}, 31.

\bibitem[Zhu et~al., 2021]{Zhu:2021tu}
Zhu, Y., Xu, Y., Liu, Q., and Wu, S. (2021).
\newblock {An Empirical Study of Graph Contrastive Learning}.
\newblock {\em arXiv.org}.

\bibitem[Zitnik and Leskovec, 2017]{zitnik2017predicting}
Zitnik, M. and Leskovec, J. (2017).
\newblock Predicting multicellular function through multi-layer tissue
  networks.
\newblock {\em Bioinformatics}, 33(14):i190--i198.

\end{thebibliography}

\end{document}